\documentclass[runningheads]{llncs}

\usepackage{eccv}

\usepackage{eccvabbrv}

\usepackage{graphicx}
\usepackage{booktabs}

\usepackage{hyperref}
\usepackage{algorithm}
\usepackage{algpseudocode}
\usepackage{makecell}
\usepackage{adjustbox}
\usepackage{wrapfig}

\newcommand{\up}[1]{\enskip\tiny\textcolor{green!60!black}{$\uparrow$#1}}
\newcommand{\down}[1]{\enskip\tiny\textcolor{red!60!black}{$\downarrow$#1}}

\usepackage[accsupp]{axessibility}  

\usepackage{graphicx}
\usepackage{makecell}
\usepackage{adjustbox}
\usepackage{booktabs}
\usepackage[table]{xcolor}
\usepackage{wrapfig}
\usepackage{subcaption}
\definecolor{myyellow}{RGB}{255, 255, 167}
\definecolor{mygreen}{RGB}{183, 255, 153}
\definecolor{shade}{RGB}{127,127,127} 

\usepackage{xcolor}
\usepackage{pifont}
\usepackage{multirow}

\usepackage{orcidlink}

\begin{document}

\title{NoisEasier: Test-Time Noise Optimization for Text-to-Video Generation} 


\author{Yujiang Pu\inst{1}\orcidlink{0009-0000-2761-5943} \and
Yu Kong\inst{1}\orcidlink{0000-0001-6271-4082}}

\authorrunning{Y. Pu and Y. Kong}

\institute{Michigan State University, East Lansing, MI 48824, USA
\email{\{puyujian,yukong\}@msu.edu}
\\
\url{https://yujiangpu20.github.io/noiseasier/}
}

\maketitle
\begin{abstract}
Diffusion models have recently advanced text-to-video (T2V) generation, yet they still struggle with fine-grained compositional alignment, such as attribute binding, spatial relations, and object interactions. While reward-based fine-tuning improves alignment, it is susceptible to reward hacking and adapts poorly to new prompt distributions. In this work, we propose \textbf{NoisEasier}, a test-time scaling framework that improves T2V generation through differentiable reward-guided noise optimization without modifying the underlying model. By combining efficient short-step generators with a multi-objective reward formulation, NoisEasier enables stable and practical test-time optimization under realistic inference budgets. Our key insight is that jointly optimizing the entire stochastic trajectory accelerates reward convergence and improves compositional alignment over optimizing only the initial latent, with negligible additional computational and time cost. Experiments on VBench and T2V-CompBench demonstrate consistent improvements across multiple backbones, achieving over 10\% average gains on challenging dimensions such as attribute binding, object interaction, and numeracy. Overall, NoisEasier serves as both a flexible alternative and a complementary enhancement to reward-based fine-tuning, establishing test-time scaling as an effective paradigm for controllable text-to-video generation.
\keywords{Video generation \and Test-time scaling \and Noise optimization}
\end{abstract}    
\section{Introduction}
\label{sec:intro}

\begin{figure*}[ht]
  \centering
  \includegraphics[width=\textwidth]{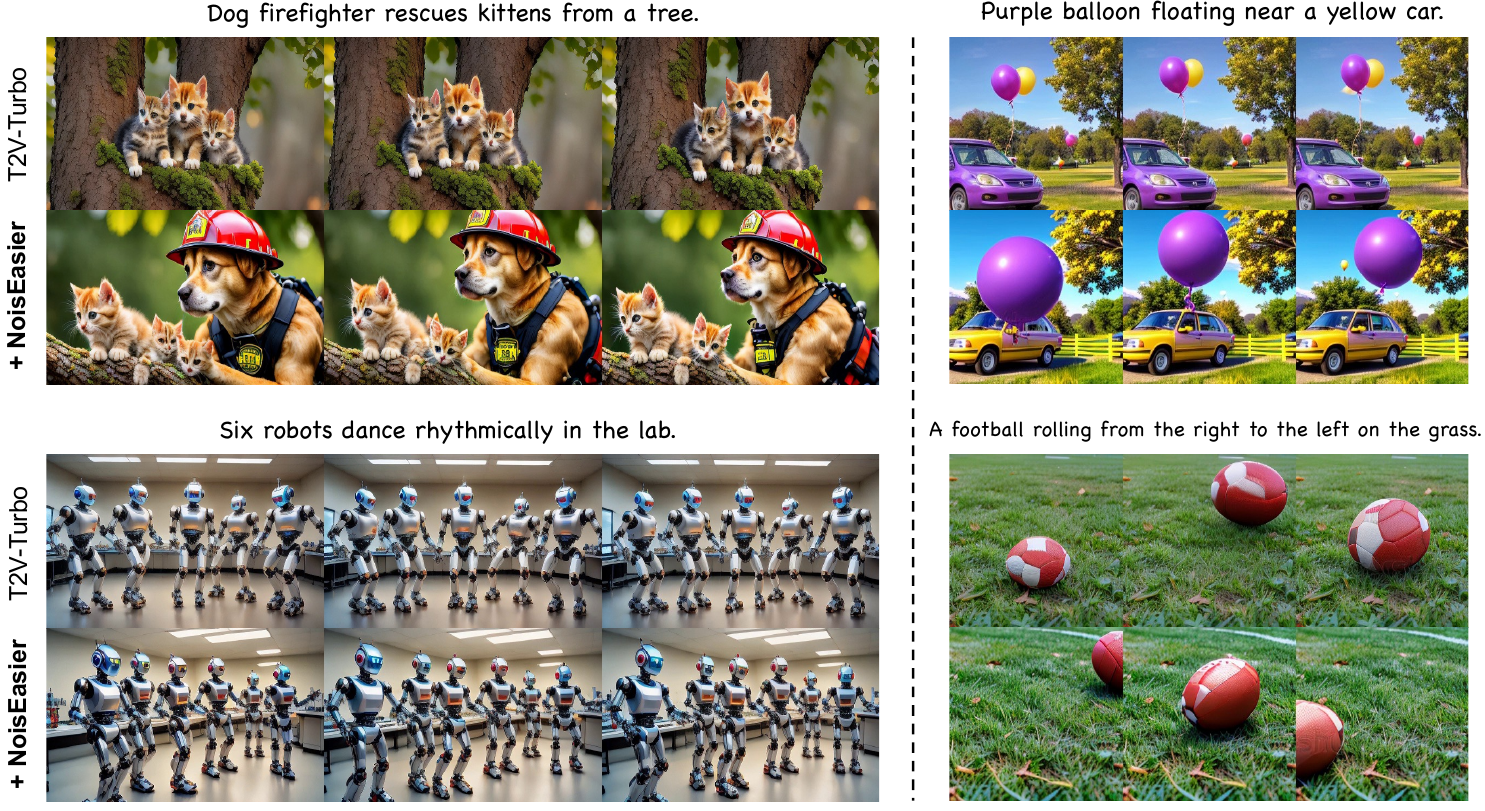}
  \caption{NoisEasier significantly boosts semantic alignment while maintaining the visual fidelity of generated videos. The results are obtained with 25-step noise optimization.
  }
  \label{fig:head}
\end{figure*}

Text-to-video (T2V) diffusion models~\cite{zhou2022magicvideo, guo2023animatediff, chen2024videocrafter2, qing2024hierarchical, wang2025lavie, wan2025wan, sora2024worldsimulators, google2025veo3} have recently achieved remarkable progress, producing videos with increasingly realistic appearance, diverse content, and coherent motion. Despite these advances, today's systems still struggle with fine-grained compositional control that faithfully reflects user intent~\cite{tian2024videotetris, sun2025t2v}.
Subtle details such as attribute bindings, object interactions, and motion dynamics often become unreliable when prompts require precise coordination among multiple semantic concepts.
These limitations reveal a broader challenge: how can we align video generation more closely with user-specified semantics while preserving visual fidelity?

Most existing works draw inspiration from Reinforcement Learning from Human Feedback (RLHF)~\cite{christiano2017deep, ouyang2022training} and adapt similar paradigms to T2V generation.
Early methods~\cite{yuan2024instructvideo, prabhudesai2024video} optimize diffusion models using differentiable rewards~\cite{kirstain2023pick, xu2023imagereward, wu2023human}, while subsequent works rely on human-preference datasets for direct preference optimization~\cite{liu2024videodpo, ding2025pami, cheng2025discriminator} or reward-based fine-tuning~\cite{wu2024boosting, liu2025improving}.
While effective in improving global alignment, these methods inherit two intrinsic limitations.
First, the learned preferences are baked into the model parameters, making continual adaptation to new reward functions or prompt distributions expensive. Second, these methods remain susceptible to reward hacking, where optimizing imperfect reward functions can degrade visual quality or motion realism.
This motivates a natural question: \emph{can we improve compositional alignment at test time without modifying the underlying video generator?}

Recent advances in test-time scaling have shown that increasing inference-time computation can substantially improve model capability without retraining.
For diffusion models, latent noise has emerged as an effective optimization variable~\cite{li2024enhancing, zhou2025golden, qi2024not, eyring2024reno, xu2025good}, leading to successful applications in image~\cite{chen2024tino, guo2024initno}, motion~\cite{karunratanakul2024optimizing}, and music generation~\cite{novack2024ditto2}.
However, extending noise optimization to T2V remains largely unexplored due to two fundamental challenges. 
First, modern T2V models~\cite{kong2024hunyuanvideo, wan2025wan} require long denoising chains and high memory consumption, making iterative gradient-based refinement prohibitively expensive. Second, video reward signals are considerably more fragile than image rewards~\cite{yuan2024instructvideo, prabhudesai2024video}, making optimization prone to instability and reward hacking.

In this work, we propose \textbf{NoisEasier}, a practical test-time scaling framework for T2V generation.
To overcome the prohibitive cost of iterative optimization, we build upon recent Video Consistency Models (VCMs)~\cite{wang2024animatelcm, li2024t2v, li2024t2v2}, which distill long diffusion sampling into only a few denoising steps, making end-to-end gradient-based refinement practical.
While VCMs address the computational bottleneck, stable optimization still requires reliable reward supervision. We therefore formulate test-time refinement as a robust multi-objective optimization problem that combines complementary semantic, aesthetic, and motion-aware rewards to mitigate reward hacking. To further strengthen compositional supervision, we introduce a lightweight negative-aware reward calibration strategy that calibrates absolute reward scores using semantically hard negative prompts, yielding more discriminative feedback.

Interestingly, the intermediate stochastic perturbations exposed by VCMs provide additional optimization variables beyond the initial latent noise. Rather than optimizing only the initialization~\cite{eyring2024reno, guo2024initno, tang2024inference, xu2025good}, we find that jointly refining all available stochastic variables consistently yields faster reward convergence and stronger compositional alignment, while introducing negligible additional runtime or memory overhead. This observation suggests that optimizing the entire denoising process provides finer control over video generation than optimizing only its initialization, offering a simple yet effective test-time scaling strategy for modern T2V models.

Extensive experiments on VBench~\cite{huang2024vbench} and T2V-CompBench~\cite{sun2024t2v}  demonstrate that NoisEasier significantly improves the performance of VCM baselines~\cite{li2024t2v, wang2024animatelcm} with only modest additional inference overhead. Compared with optimizing the initial latent alone, jointly optimizing the intermediate perturbations yields more consistent improvements on compositional dimensions, including attribute binding, spatial relations, and object interactions. 
Furthermore, NoisEasier continues to improve reward-finetuned models, suggesting that offline fine-tuning does not fully amortize reward signals into the model parameters. 
Consequently, NoisEasier serves not only as a flexible alternative to reward fine-tuning, but also as a complementary test-time scaling mechanism that unlocks additional gains beyond offline alignment.

In summary, our main contributions are as follows:
\begin{itemize}
    \item We propose \textbf{NoisEasier}, a test-time scaling framework for text-to-video generation that enables efficient and robust reward-guided optimization without model fine-tuning.

    \item We develop an efficient optimization pipeline based on short-step Video Consistency Models together with a robust reward formulation for stable compositional optimization.

    \item We show that jointly optimizing the intermediate perturbations consistently accelerates reward convergence and improves compositional alignment over initial-noise optimization with negligible additional computational cost.
\end{itemize}
\section{Related Work}
\label{sec:related}

\textbf{Video Generation with Human Feedback.}
Reinforcement Learning from Human Feedback (RLHF)~\cite{christiano2017deep, ouyang2022training} has shown great success in aligning large language models (LLMs) with human intent, motivating its extension to image and video generation. Existing work typically follows two paradigms. One is \emph{preference-driven optimization}, where models are trained directly on large-scale human preference datasets. This includes policy gradient methods like DDPO~\cite{black2023training} and DPOK~\cite{fan2023dpok}, as well as direct preference optimization methods~\cite{liang2024rich, wallace2024diffusion, yang2024using, liu2024videodpo} that bypass explicit reward models. While effective, these approaches require large-scale preference annotations and cannot easily incorporate structured human knowledge. The other is \emph{reward-driven optimization}, where pretrained reward models provide differentiable feedback for fine-tuning generative models~\cite{yuan2024instructvideo, prabhudesai2024video} or guide distillation for faster inference~\cite{li2024reward, li2024t2v, li2024t2v2}. Although these methods reduce training cost and improve stability, they are generally one-shot systems that lack mechanisms for iterative refinement. 
In contrast, we directly optimize latent noise via differentiable rewards, enabling preference-controllable text-to-video generation with efficient test-time refinement.

\noindent\textbf{Test-time Noise Optimization.}
Recent studies show that the initial latent noise strongly influences diffusion generation quality, with certain \emph{golden noise} consistently yielding outputs that are more faithful and aesthetically pleasing. Building on this observation, test-time methods that operate on noise can be broadly categorized into two directions: \emph{selection-based noise search} and \emph{optimization-based noise refinement}. Selection-based methods~\cite{li2024enhancing, zhou2025golden, xu2025good} treat sampling as a search problem over initial noises or trajectories and select the best sample under a fixed backbone. In contrast, optimization-based methods directly refine the noisy latent via gradient-based updates during sampling. DOODL~\cite{wallace2023end} pioneered end-to-end differentiable latent optimization, and subsequent work extended this paradigm to images~\cite{chen2024tino, guo2024initno, tang2024inference, zhou2025golden} and other modalities, including music~\cite{novack2024ditto} and motion~\cite{karunratanakul2024optimizing}. While effective, optimization-based approaches can be prohibitively slow at inference time; for example, DOODL~\cite{wallace2023end} and D-Flow~\cite{ben2024d} may take tens of minutes per sample. ReNO~\cite{eyring2024reno} partially mitigates this bottleneck by applying optimization to one-step diffusion models, reducing runtime to tens of seconds. In this work, we revisit noise optimization for text-to-video generation, which remains underexplored due to the high cost of video sampling and the scarcity of suitable spatiotemporal reward signals.
\section{Method}

\subsection{Problem Formulation}

Let $\mathcal{G}_\theta$ denote a pretrained text-to-video (T2V) diffusion model. Given a text prompt $c$, the generation process starts from an initial latent noise $\mathbf{z}_T \sim \mathcal{N}(0,\mathbf{I})$ and progressively denoises it into a video sequence. Existing test-time optimization methods formulate noise refinement by directly optimizing the initial latent:
\begin{equation}
\mathbf{z}_T^*=\arg\max_{\mathbf{z}_T}\;
\mathcal{R}(\mathcal{G}_\theta(\mathbf{z}_T,c)),
\label{eq:obj}
\end{equation}
where $\mathcal{R}(\cdot)$ denotes a differentiable reward measuring semantic alignment or perceptual quality.
While conceptually simple, directly optimizing $\mathbf{z}_T$ for T2V generation is computationally expensive, as gradients must propagate through dozens of denoising steps. Previous works~\cite{clark2023directly, prabhudesai2023aligning, prabhudesai2024video} alleviate this cost through gradient truncation or partial backpropagation, sacrificing end-to-end optimization while remaining computationally demanding.
In contrast, we build upon recent Video Consistency Models (VCMs)~\cite{wang2024animatelcm, li2024t2v}, which distill long diffusion sampling into only a few denoising steps. 

Specifically, a consistency model learns a mapping that directly predicts the clean sample along the probability-flow ODE trajectory:
$f: (\mathbf{z}_t, t, c) \mapsto \mathbf{z}_\kappa,  t \in [\kappa, T]$, where $\kappa$ is a small positive constant. The model is trained with a self-consistency objective that enforces identical predictions for different noisy states originating from the same trajectory:
\begin{equation}
f(\mathbf z_t,t,c)=f(\mathbf z_{t'},t',c),
\quad
\forall\,t,t'\in[\kappa,T],
\label{eq:self_consistency}
\end{equation}
allowing the original diffusion trajectory to be approximated using only a small number of denoising steps. This substantially reduces the cost of gradient propagation and enables practical end-to-end optimization for video generation.
During inference, VCM alternates between predicting the clean sample
$\hat{\mathbf z}_0=f(\mathbf z_t,t,c)$
and injecting Gaussian perturbations
$\boldsymbol\varepsilon_t\sim\mathcal N(0,\mathbf I)$
to obtain the next latent
\begin{equation}
\mathbf z_{t-1}
=
\sqrt{\bar\alpha_{t-1}}\,
\hat{\mathbf z}_0
+
\sqrt{1-\bar\alpha_{t-1}}\,
\boldsymbol\varepsilon_t.
\label{eq:vcm_step}
\end{equation}

Unlike conventional diffusion sampling, the stochastic formulation of VCM naturally exposes the injected perturbations as additional optimization variables. Rather than optimizing only the initial latent as in Eq.~\ref{eq:obj}, we jointly optimize all available stochastic variables throughout the denoising process:
\begin{equation}
(\mathbf z_T^*,\{\boldsymbol\varepsilon_t^*\})
=
\arg\max_{\mathbf z_T,\{\boldsymbol\varepsilon_t\}}
\mathcal R\!\left(
\mathcal G_\theta(\mathbf z_T,c;
\{\boldsymbol\varepsilon_t\})
\right).
\end{equation}

This formulation provides finer control over the denoising trajectory and consistently accelerates reward convergence compared with optimizing only the initial latent, while introducing negligible additional computational overhead.

\begin{figure*}[t]
    \centering
    \includegraphics[width=\linewidth]{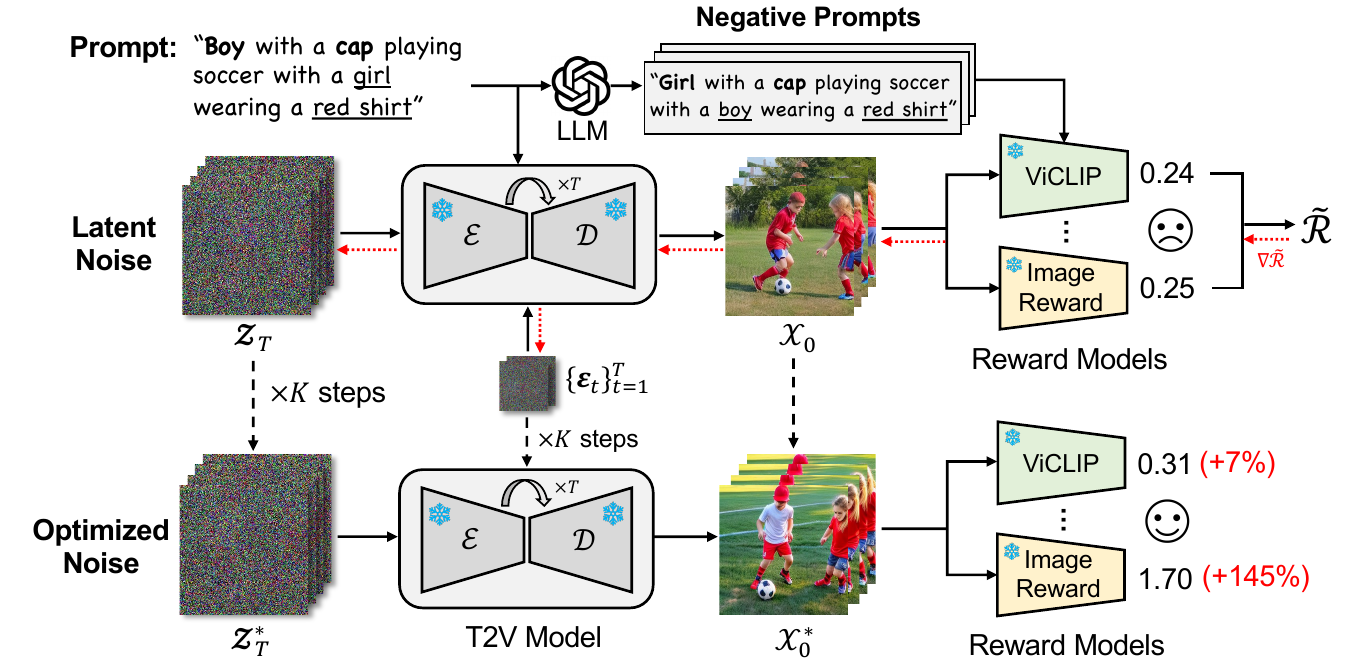}
    \caption{Overview of the proposed NoisEasier framework. Given a text prompt, both the initial latent noise and intermediate perturbations are iteratively optimized using gradients from multiple reward models. Negative prompts generated by an LLM further calibrate the reward signal for fine-grained compositional feedback.}
    \label{fig:pipeline}
\end{figure*}

\subsection{Reward Models for Video Generation}

While human-preference reward models are well studied for image generation~\cite{kirstain2023pick, xu2023imagereward, wu2023human2}, reward modeling for video remains underexplored.
Recent works such as VideoRM~\cite{wu2024boosting}, VideoScore~\cite{he2024videoscore}, and VisionReward~\cite{xu2024visionreward} leverage multimodal large language models (MLLMs) to assess generated videos.
However, directly using large MLLM-based reward models inside gradient-based noise refinement can be computationally prohibitive, since the reward must be evaluated repeatedly during optimization.
Therefore, we focus on lightweight, differentiable rewards built on pretrained vision-language models (VLMs), which provide strong semantic alignment signals while remaining efficient for test-time optimization. Specifically, we consider three types of reward objectives:

\noindent\textbf{Image-Text Alignment Reward.} We employ HPSv2~\cite{wu2023human2} and ImageReward~\cite{xu2023imagereward}, both trained to model human preferences for image-text relevance. HPSv2 uses a CLIP-based~\cite{radford2021learning} architecture with a fine-tuned reward head, while ImageReward adopts BLIP~\cite{li2022blip} as its backbone, fine-tuned with supervised human feedback to improve perceptual alignment. Both models produce scalar scores that reflect the semantic alignment between the generated image and the text prompt. In practice, we compute the reward by evaluating each frame-text pair and take the mean score across all frames as the final image-text alignment reward.

\noindent\textbf{Video-Text Alignment Reward.} Image-level reward models provide strong semantic cues but lack temporal perception, making them ineffective at capturing motion consistency in video generation. To address this, we incorporate ViCLIP~\cite{wang2023internvid}, a video-language model trained on the large-scale InternVid dataset. ViCLIP jointly encodes short video clips and text descriptions to assess temporal and semantic consistency. In practice, we sample two temporally offset clips from the generated video, each consisting of 8 frames, and compute their similarity to the text prompt. The final reward is the average of the two clip-text similarity scores, which provides video-level alignment feedback.

\noindent\textbf{Motion-aware Reward.} 
Although VLM-based rewards help improve semantic alignment with the prompt, they often push the model to favor static content to maximize alignment scores. As a result, the generated videos may lack meaningful motion or introduce temporal artifacts. To mitigate this, we introduce a motion-aware reward that explicitly encourages localized dynamics while regularizing temporal coherence.
Specifically, we use RAFT~\cite{teed2020raft} to estimate optical flow $\{\mathbf{F}_t\}_{t=1}^{T-1}$, where $\mathbf{F}_t \in \mathbb{R}^{2\times H \times W}$ captures motion from frame $t$ to $t+1$. 
We define the spatial mean flow vector $\bar{\mathbf{f}}_t = \frac{1}{HW}\sum_{h,w}\mathbf{F}_t[:,h,w]$ and subtract it to suppress global camera motion.
The motion reward is computed as:
\begin{equation}
\mathcal{R}_{\text{mo}}
=
\tanh \Bigl(
\tfrac{1}{T-1}\sum_{t=1}^{T-1}\lVert \mathbf{F}_t-\bar{\mathbf{f}}_t\rVert_1
\Bigr)
+
\lambda \Bigl(
1-\tanh\Bigl(\tfrac{1}{T-2}\sum_{t=1}^{T-2}\lVert \mathbf{F}_{t+1}-\mathbf{F}_t\rVert_1\Bigr)
\Bigr),
\label{eq:motion}
\end{equation}
where the first term encourages localized motion beyond global translation, and the second term penalizes abrupt changes in flow magnitude to promote temporally coherent dynamics. In practice, this objective effectively mitigates the static bias induced by image-based rewards, leading to more expressive and dynamic video generation.

\subsection{Negative-Aware Reward Calibration}

Most existing reward models~\cite{kirstain2023pick, wu2023human, wu2023human2, xu2023imagereward} are trained on human preferences using pairwise ranking, providing only weak supervision for fine-grained compositional reasoning.
Prior studies~\cite{yuksekgonul2022and, doveh2023dense, trager2023linear} further show that VLM-based evaluators often behave like \emph{bag-of-words} classifiers.
Consequently, reward signals can be overly coarse and insensitive to subtle semantic errors.
To address this limitation, we introduce \emph{negative-aware} reward calibration, which contrasts the target prompt against semantically perturbed distractors rather than relying solely on absolute reward scores.

Given a prompt $c^{+}$, we utilize a large language model (LLM) to construct a set of hard negative prompts $\{c^-_j\}_{j=1}^{N}$ that are syntactically similar but semantically incorrect, as illustrated in Figure~\ref{fig:pipeline}. 
These negatives encourage the reward to capture fine-grained semantic distinctions rather than superficial lexical similarity. 
Given a generated video $x$ and a reward model $\mathcal{R}(\cdot)$ for a video--text pair, the calibrated reward is defined as
\begin{equation}
\tilde{\mathcal{R}} 
= \mathcal{R}(x, c^{+})
  - \tau \log \sum_{j=1}^{N} 
    \exp\left(
      \frac{\mathcal{R}(x, c^-_{j}) - \mathcal{R}(x, c^{+}) + m}{\tau}
    \right),
\end{equation}
where \(m\) is a margin enforcing separation between the target prompt and its negatives, and \(\tau\) is a temperature parameter controlling the sharpness of the contrastive penalty. Intuitively, the calibrated reward encourages the generated video to achieve a high score for the target prompt while suppressing semantically similar distractors, making the optimization process more sensitive to compositional errors such as attribute binding, spatial relations, and numeracy.

Finally, we optimize both the initial latent noise $\mathbf{z}_T$ and the stochastic perturbations $\{\boldsymbol{\varepsilon}_t\}_{t=1}^{T}$ through gradient ascent on a weighted combination of the calibrated rewards.
For brevity, we denote all optimization variables by
$\boldsymbol{\epsilon}=\{\mathbf{z}_T,\boldsymbol{\varepsilon}_1,\ldots,\boldsymbol{\varepsilon}_T\}$,
leading to the update rule
\begin{equation}
\boldsymbol{\epsilon}^{k+1} = \boldsymbol{\epsilon}^k +
\eta \cdot \nabla_{\boldsymbol{\epsilon}^k} \left[
\sum_i \alpha_i \tilde{\mathcal{R}}_i(\mathcal{G}_\theta(\boldsymbol{\epsilon}^k,\, c), c)
\right],
\label{eq:update-rule}
\end{equation}
where $\eta$ denotes the learning rate and $\alpha_i$ are the weights of individual reward components.
Together, the proposed reward formulation and joint stochastic optimization enable stable and effective test-time refinement.

\section{Experiment}
\label{sec:experiment}

\renewcommand{\arraystretch}{1.08} 
\setlength{\tabcolsep}{3pt} 
\begin{table}[tb]
  \caption{Quantitative evaluation on VBench (\%). The best results among open-source models are in \textbf{bold} and the second best is \underline{underlined}. * denotes our reproduction.}
  \label{tab:vbench}
  \centering
  \resizebox{\linewidth}{!}{
  \begin{tabular}{lllllll} 
\toprule
Models & \makecell{Overall \\ Consist.} & \makecell{Multiple \\ Objects} & 
\makecell{Color} & 
\makecell{Spatial \\ Relation.} &  \makecell{Motion \\ Smooth.} & \makecell{Dynamic \\ Degree} \\
\midrule
\textcolor{shade}{\emph{(Closed-source)}} &  &  & &  &  &  \\
\textcolor{shade}{Gen-3}~\cite{runway2024gen3}   & \textcolor{shade}{26.69} & \textcolor{shade}{53.64} & \textcolor{shade}{80.90} & \textcolor{shade}{65.09} & \textcolor{shade}{99.23} & \textcolor{shade}{60.14} \\
\textcolor{shade}{Sora}~\cite{sora2024worldsimulators}     & \textcolor{shade}{26.26} & \textcolor{shade}{70.85} & \textcolor{shade}{80.11} & \textcolor{shade}{74.29} & \textcolor{shade}{98.74} & \textcolor{shade}{79.91} \\
\textcolor{shade}{Veo 3}~\cite{google2025veo3}    & \textcolor{shade}{27.88} & \textcolor{shade}{82.20} & \textcolor{shade}{82.48} & \textcolor{shade}{84.26} & \textcolor{shade}{99.16} & \textcolor{shade}{72.43} \\
\midrule
\textcolor{shade}{\emph{(Open-source)}} &  &  & &  &  &  \\
ModelScope~\cite{wang2023modelscope}      & 25.67 & 38.98 & 81.72 & 33.68 & 95.79 & 66.39 \\
VideoCrafter2~\cite{chen2024videocrafter2}   & 28.23 & 40.66 & 92.92 & 35.86 & 97.73 & 42.50 \\
CogVideoX-2B~\cite{yang2024cogvideox}   & 27.33 & 66.59 & 83.01 & \textbf{74.27} & 97.51 & 64.79 \\
Vchitect-2.0~\cite{fan2025vchitect}    & 27.57 & 68.84 & 87.04 & \underline{57.55} & \textbf{98.98} & 63.89 \\
\midrule
AnimateLCM*~\cite{wang2024animatelcm}    & 25.74 & 30.78 & 81.84 & 39.94 & 98.19 & 29.17 \\
\rowcolor{gray!15}
\textbf{+ NoisEasier (Ours)} & 28.79\up{3.05} & 44.04\up{13.26} & 88.22\up{6.38} & 46.70\up{6.76} & 98.57\up{0.38} & 30.56\up{1.39} \\
\midrule
T2V-Turbo (MS)~\cite{li2024t2v}    & 27.51 & 58.63 & 89.67 & 45.74 & 95.64 & 66.39 \\
T2V-Turbo (MS)*   & 27.40 & 54.50 & 89.90 & 46.62 & 95.51 & \underline{70.83} \\
\rowcolor{gray!15}
\textbf{+ NoisEasier (Ours)} & \textbf{32.05}\up{4.65} & \underline{76.62}\up{22.12} & \textbf{94.38}\up{4.48} & 48.93\up{2.31} & 95.29\down{0.22} & \textbf{71.39}\up{0.56} \\
\midrule
T2V-Turbo (VC2)~\cite{li2024t2v}   & 28.16 & 54.65 & 89.90 & 38.67 & 97.34 & 49.17 \\
T2V-Turbo (VC2)*  & 28.12 & 54.33 & 90.59 & 39.95 & 96.29 & 51.67 \\
\rowcolor{gray!15}
\textbf{+ NoisEasier (Ours)} & \underline{31.95}\up{3.83} & \textbf{79.18}\up{24.85} & \underline{93.63}\up{3.04} & 52.61\up{12.66} & 96.10\down{0.19} & 59.17\up{7.50} \\
\bottomrule
\end{tabular}
} 
\end{table}

\noindent \textbf{Experimental Setup.}
We evaluate NoisEasier on two VCM backbones: AnimateLCM~\cite{wang2024animatelcm} and T2V-Turbo~\cite{li2024t2v}.
For T2V-Turbo, we consider two distilled variants: T2V-Turbo (MS) and T2V-Turbo (VC2), which are distilled from ModelScope~\cite{wang2023modelscope} and VideoCrafter2~\cite{chen2024videocrafter2}, respectively.
T2V-Turbo (MS) generates videos at $256\times256$ resolution, while T2V-Turbo (VC2) produces $512\times320$ videos. 
AnimateLCM generates videos at $512\times512$ resolution. All three backbones support fast sampling with 4–8 denoising steps. To balance generation quality and efficiency, we adopt a 4-step sampler for fast inference. For noise refinement, we optimize the latent variables for 25 iterations using AdamW with gradient clipping and a learning rate of 0.01. 
Unless otherwise specified, all experiments use the same optimization configuration across backbones.
To make noise optimization feasible, we employ gradient checkpointing~\cite{chen2016training} together with mixed-precision inference to reduce memory overhead. All the experiments are conducted on two RTX 6000 Ada GPUs.
More implementation details can be found in the supplementary material.

\noindent\textbf{Benchmarks and Metrics.} We evaluate NoisEasier on two widely used benchmarks: VBench~\cite{huang2024vbench} and T2V-CompBench~\cite{sun2025t2v}. 
VBench provides a comprehensive evaluation across 16 dimensions covering visual quality and semantic alignment. 
Following common practice, we report results on four semantic dimensions (\emph{overall consistency}, \emph{multiple objects}, \emph{color}, and \emph{spatial relations}) and two motion-related dimensions, i.e., \emph{motion smoothness} and \emph{dynamic degree}. 
T2V-CompBench focuses on compositional reasoning in video generation and evaluates seven categories, including \emph{consistent/dynamic attribute binding}, \emph{spatial relationships}, \emph{action/motion binding} \emph{object interactions}, and \emph{numeracy}.

\begin{table}[t]
\centering
\caption{Quantitative evaluation on T2V-CompBench.
The best results among open-source models are in \textbf{bold}, and the second best are \underline{underlined}.
* denotes reproduction.}
\label{tab:compbench}
\renewcommand{\arraystretch}{1.18}
\resizebox{\textwidth}{!}{%
\begin{tabular}{llllllll}
\toprule
Models &
\makecell{Consist\\ Attr.} &
\makecell{Dynamic \\Attr.} &
Spatial &
Motion &
Action &
Interaction &
Numeracy \\
\midrule
\multicolumn{8}{l}{\textcolor{shade}{\emph{(Closed-source)}}} \\
\textcolor{shade}{Gen-3}~\cite{runway2024gen3} & \textcolor{shade}{0.5980} & \textcolor{shade}{0.0687} & \textcolor{shade}{0.5194} & \textcolor{shade}{0.2754} & \textcolor{shade}{0.5233} & \textcolor{shade}{0.5906} & \textcolor{shade}{0.2306} \\
\textcolor{shade}{Dreamina}~\cite{capcut2024dreamina} & \textcolor{shade}{0.6913} & \textcolor{shade}{0.0051} & \textcolor{shade}{0.5773} & \textcolor{shade}{0.2361} & \textcolor{shade}{0.5924} & \textcolor{shade}{0.6824} & \textcolor{shade}{0.4380} \\
\textcolor{shade}{PixVerse}~\cite{pixverse2024} & \textcolor{shade}{0.7060} & \textcolor{shade}{0.0624} & \textcolor{shade}{0.5979} & \textcolor{shade}{0.2867} & \textcolor{shade}{0.8722} & \textcolor{shade}{0.8309} & \textcolor{shade}{0.6066} \\
\textcolor{shade}{Kling}~\cite{kuaishou2024kling} & \textcolor{shade}{0.6931} & \textcolor{shade}{0.0098} & \textcolor{shade}{0.5690} & \textcolor{shade}{0.2562} & \textcolor{shade}{0.5787} & \textcolor{shade}{0.7128} & \textcolor{shade}{0.4413} \\
\midrule
\multicolumn{8}{l}{\emph{(Open-source)}} \\
ModelScope~\cite{wang2023modelscope} & 0.5148 & 0.0161 & 0.4118 & 0.2408 & 0.3639 & 0.4613 & 0.1986 \\
Show-1~\cite{zhang2024show} & 0.5670 & 0.0115 & 0.4544 & 0.2291 & 0.3881 & 0.6244 & 0.3086 \\
VideoTetris~\cite{tian2024videotetris} & 0.6211 & 0.0104 & 0.4832 & 0.2249 & 0.4939 & 0.6578 & 0.3467 \\
VideoCrafter2~\cite{chen2024videocrafter2} & 0.6182 & 0.0103 & 0.4838 & 0.2259 & 0.5030 & 0.6365 & 0.3330 \\
Open-Sora 1.2~\cite{zheng2024open} & 0.5639 & \underline{0.0189} & 0.5063 & 0.2468 & 0.4833 & 0.5039 & 0.3719 \\
CogVideoX-5B~\cite{yang2024cogvideox} & 0.6164 & \textbf{0.0219} & 0.5172 & \textbf{0.2658} & 0.5333 & 0.6069 & 0.3706 \\
\midrule
AnimateLCM*~\cite{wang2024animatelcm} & 0.6746 & 0.0096 &  0.4651 &  0.2208 &  0.3425 & 0.4504 & 0.2494 \\
\rowcolor{gray!15}
\textbf{+ NoisEasier (Ours)} & 0.7807\up{0.1061} & 0.0183\up{0.0087} & 0.5113\up{0.0462} & 0.2281\up{0.0073} & 0.4362\up{0.0937} & 0.5365\up{0.0861} & 0.2567\up{0.0073} \\
\midrule
T2V-Turbo (MS)*~\cite{li2024t2v} & 0.7313 & 0.0124 & 0.4620 & 0.2306 & 0.4923 & 0.5822 & 0.3030 \\
\rowcolor{gray!15}
\textbf{+ NoisEasier (Ours)} & \underline{0.8375}\up{0.1062} & 0.0147\up{0.0023} & \underline{0.5259}\up{0.0639} & \underline{0.2502}\up{0.0196} & \underline{0.6477}\up{0.1554} & 0.6743\up{0.0921} & \textbf{0.3983}\up{0.0953} \\
\midrule
T2V-Turbo (VC2)*~\cite{li2024t2v} & 0.7852 & 0.0113 & 0.5079 & 0.2222 & 0.6241 & \underline{0.7017} & 0.2900 \\
\rowcolor{gray!15}
\textbf{+ NoisEasier (Ours)} & \textbf{0.8737}\up{0.0885} & 0.0097\down{0.0016} & \textbf{0.5660}\up{0.0581} & 0.2343\up{0.0121} & \textbf{0.7524}\up{0.1283} & \textbf{0.7915}\up{0.0898} & \underline{0.3753}\up{0.0853} \\
\bottomrule
\end{tabular}%
}
\end{table}

\subsection{Main Results}
We first report quantitative results on VBench, comparing against both closed-source models and open-source baselines. As shown in Table~\ref{tab:vbench}, NoisEasier consistently improves VCM-based generators across different backbones. 
On AnimateLCM, NoisEasier increases \emph{Overall Consistency} from 25.74\% to 28.79\%, with notable gains in compositional dimensions such as \emph{Multiple Objects} and \emph{Spatial Relation}. Similar trends are observed on T2V-Turbo, where the improvements are even more pronounced. For example, \emph{Multiple Objects} increases from 54.50\% to 76.62\% on T2V-Turbo (MS) and from 54.33\% to 79.18\% on T2V-Turbo (VC2), while \emph{Spatial Relation} improves by 12.66\% on the VC2 variant. Although \emph{Motion Smoothness} decreases slightly for the T2V-Turbo variants, the consistent increase in \emph{Dynamic Degree} suggests that NoisEasier produces richer motion dynamics while largely preserving temporal coherence.

\begin{figure*}[!ht]
  \centering
  \includegraphics[width=\textwidth]{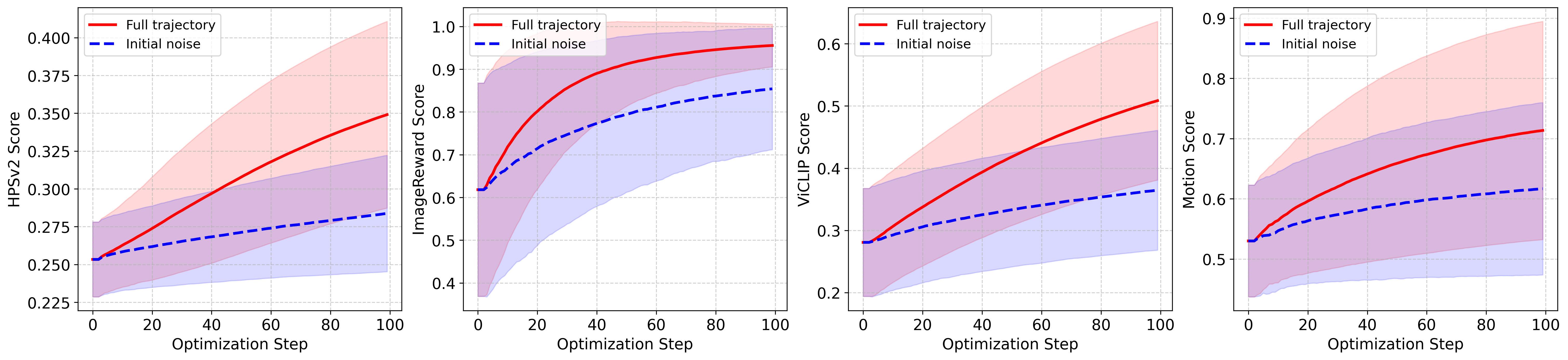}
  \caption{Reward curve during noise optimization. The red solid line represents full noise optimization, while the blue dotted line denotes initial noise optimization. All scores are scale-normalized.}
  \label{fig:reward-curve}
\end{figure*}

 NoisEasier also yields consistent gains on T2V-CompBench across diverse compositional dimensions, as shown in Table~\ref{tab:compbench}. On AnimateLCM, it substantially improves \emph{Attribute/Action Binding} while also boosting \emph{Spatial relationship} and \emph{Interaction}. Similar trends hold for T2V-Turbo, where \emph{Consist. Attribute Binding} increases to 0.8375 (MS) and 0.8737 (VC2), and \emph{Action binding} improves notably on both variants. Importantly, Numeracy also sees sizable gains (e.g., 0.0953 on MS), indicating improved robustness in reasoning about discrete object counts. These results further support the effectiveness of our negative-aware reward calibration in providing more precise compositional feedback.

\subsection{Ablation Study}

\begin{wraptable}{r}{0.54\columnwidth}
\vspace{-34pt} 
\centering
\caption{The effect of reward models for NoisEasier on VBench. The best score among individual reward models is marked in \colorbox{mygreen}{green}, while the second-best is in \colorbox{myyellow}{yellow}.}
\label{tab:ablation}
\setlength{\tabcolsep}{3.5pt}
\renewcommand{\arraystretch}{1.08}
\footnotesize
\resizebox{0.50\columnwidth}{!}{%
\begin{tabular}{lcccccc}
\toprule
Models &
\makecell{Overall\\Consist.} &
\makecell{Multiple\\Objects} & Color &
\makecell{Spatial\\Relation.} &
\makecell{Motion\\Smooth.} &
\makecell{Dynamic\\Degree} \\
\midrule
T2V-Turbo (MS)* & 27.40 & 54.50 & 89.90 & 46.62 & 95.51 & 70.83 \\
+ HPSv2         & 27.85 & 64.66 & 90.20 & \cellcolor{myyellow}46.21 & 95.19 & 54.72 \\
+ ImageReward   & \cellcolor{myyellow}28.24 & \cellcolor{mygreen}73.78 & \cellcolor{myyellow}90.27 & \cellcolor{mygreen}47.16 & \cellcolor{myyellow}95.36 & 61.11 \\
+ ViCLIP        & \cellcolor{mygreen}36.21 & \cellcolor{myyellow}68.05 & \cellcolor{mygreen}91.19 & 41.92 & 94.98 & \cellcolor{myyellow}63.89 \\
+ Motion        & 26.05 & 47.68 & 87.32 & 42.07 & \cellcolor{mygreen}95.59 & \cellcolor{mygreen}85.56 \\
\midrule
All w/o negatives & 31.74 & 75.82 & 93.29 & 47.95 & 95.38 & 73.33 \\
All (NoisEasier)  & 32.05 & 76.62 & 94.38 & 48.93 & 95.29 & 71.39 \\
\bottomrule
\end{tabular}%
}
\vspace{-10pt} 
\end{wraptable}

\begin{figure*}[t]
    \centering
    \begin{subfigure}[t]{0.38\textwidth}
        \centering
        \includegraphics[width=\textwidth]{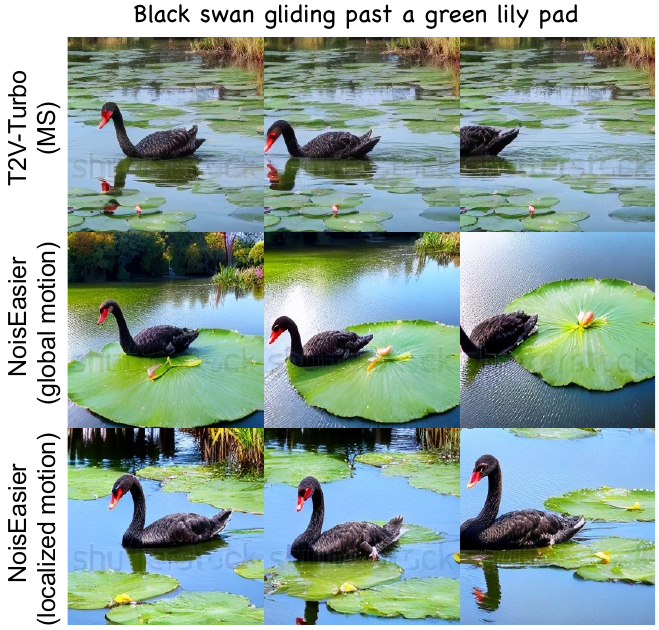}
        \caption{Global \emph{vs.} localized motion.}
        \label{fig:motion}
    \end{subfigure}
    \begin{subfigure}[t]{0.59\textwidth}
        \centering
        \includegraphics[width=\textwidth]{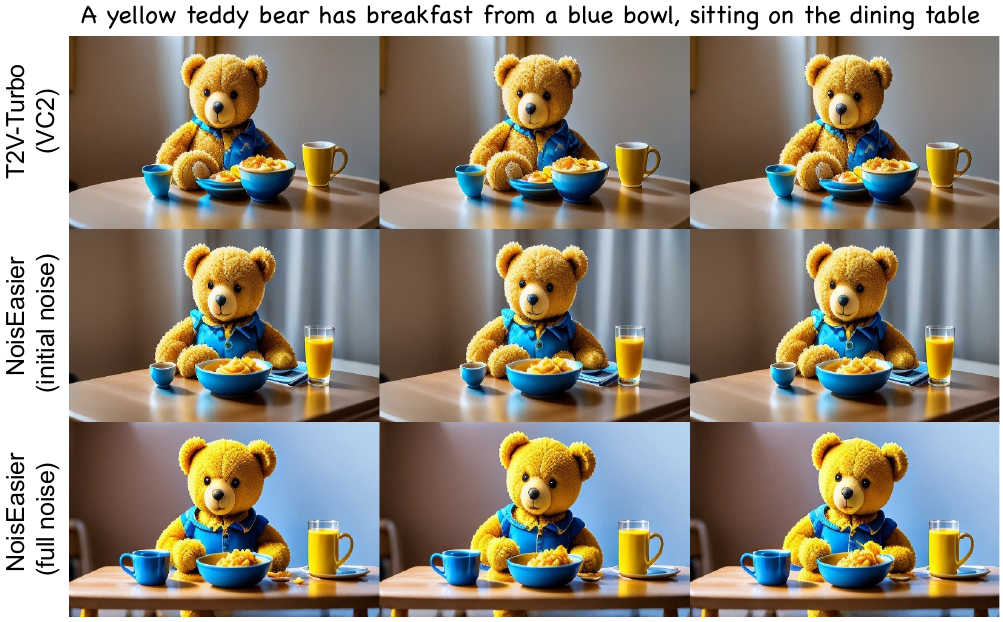}
        \caption{Initial noise optimization \emph{vs.} Full noise optimization.}
        \label{fig:noise}
    \end{subfigure}
    \caption{The effect of (a) motion reward; and (b) noise optimization formulation.}
    \label{fig:ablation}
\end{figure*}

\textbf{Contribution of each reward model.} As shown in Table~\ref{tab:ablation}, different reward models emphasize different aspects of generation. ImageReward and ViCLIP contribute most to semantic alignment, but optimizing semantic rewards alone tends to yield overly static videos, reflected by reduced \emph{Dynamic Degree}. In contrast, the motion reward alone increases temporal dynamics and motion coherence, yet degrades semantic metrics, highlighting the need to balance complementary objectives rather than relying on a single signal. Figure~\ref{fig:motion} further explains the dynamic term in our motion reward: by subtracting \emph{global motion} in Eq.~\ref{eq:motion}, the resulting \emph{localized motion} suppresses camera shifts while emphasizing subject movement. Combining all rewards yields complementary gains across semantic and motion dimensions, and negative-aware calibration further strengthens compositional alignment, with only a slight decrease in motion-related scores since it primarily reshapes VLM-based objectives.

\begin{wraptable}{r}{0.54\columnwidth}
\vspace{-1.2cm}
\centering
\caption{Initial \emph{vs}. full noise optimization.}
\label{tab:noise}
\setlength{\tabcolsep}{3.5pt}
\renewcommand{\arraystretch}{1.08}
\footnotesize
\resizebox{0.50\columnwidth}{!}{%
\begin{tabular}{lcccccc}
\toprule
Models &
\makecell{Overall\\Consist.} &
\makecell{Multiple\\Objects} & Color &
\makecell{Spatial\\Relation.} &
\makecell{Motion\\Smooth.} &
\makecell{Dynamic\\Degree} \\
\midrule
AnimateLCM+Init. & 26.76 & 43.28 & 85.61 & 44.52 & 98.36 & 30.83 \\
\rowcolor{gray!15}AnimateLCM+Full. & 28.79 & 44.04 & 88.22 & 46.70 & 98.57 & 30.56 \\
\hline
T2V-Turbo (MS)+Init. & 29.67 & 72.58 & 90.03 & 44.10 & 95.58 & 66.94 \\
\rowcolor{gray!15}T2V-Turbo (MS)+Full.    & 32.05 & 76.62 & 94.38 & 48.93 & 95.29 & 71.39 \\
\hline
T2V-Turbo (VC2)+Init. & 29.97 & 75.41 & 92.98 & 45.15 & 96.16 & 60.83 \\
\rowcolor{gray!15}T2V-Turbo (VC2)+Full. & 31.95 & 79.18 & 93.63 & 52.61 & 96.10 & 59.17 \\
\bottomrule
\end{tabular}%
}
\vspace{-20pt}
\end{wraptable}
\noindent\textbf{Initial noise \emph{vs.} Full trajectory optimization.} 
To quantify the benefit of full noise optimization, we compare against a variant that updates only the initial latent noise. As shown in Table~\ref{tab:noise}, 
full-trajectory optimization provides additional control beyond initial-noise-only refinement, with the largest gains on compositional metrics such as color and spatial relations. Interestingly, motion-related dimensions show smaller gains or occasional advantages for initial noise optimization, suggesting lower sensitivity to intermediate-noise refinement.
More results on T2V-CompBench can be found in Table~\ref{tab:compbench-abl}. 
Figure~\ref{fig:reward-curve} further demonstrates that jointly optimizing the full noise trajectory accelerates reward convergence.
We also provide a qualitative example in Figure~\ref{fig:noise}: the initial noise mainly controls global appearance and coarse layout, whereas intermediate perturbations refine finer structures and stylistic details, resulting in sharper contrast and more visually pleasing outputs. Overall, these results validate the effectiveness of full trajectory optimization for improving both quantitative performance and visual quality.

\begin{wraptable}{r}{0.54\columnwidth}
\vspace{-1.2cm}
\centering
\caption{Effect of different negative prompt construction.}
\label{tab:prompt}
\setlength{\tabcolsep}{3.5pt}
\renewcommand{\arraystretch}{1.08}
\footnotesize
\resizebox{0.50\columnwidth}{!}{%
\begin{tabular}{lcccccc}
\toprule
Models &
\makecell{Overall\\Consist.} &
\makecell{Multiple\\Objects} & Color &
\makecell{Spatial\\Relation.} &
\makecell{Motion\\Smooth.} &
\makecell{Dynamic\\Degree} \\
\midrule
NoisEaiser w/o neg. & 31.74 & 75.82 & 93.29 & 47.95 & 95.38 & 73.33 \\
\midrule
Random sample  &  31.83 &  75.41 &  93.14 &  46.39 &  95.46 &  71.11 \\
Gemini-2.5 Flash & 32.00 & 76.55 & 93.81 & 48.72 & 95.53 & 74.44 \\
GPT-4o (Ours) &  32.05 & 76.62 & 94.38 & 48.93 & 95.29 & 71.39 \\
\bottomrule
\end{tabular}%
}
\vspace{-10pt}
\end{wraptable}

\noindent \textbf{Comparison of different negative prompts.}
We compare LLM-generated negatives against randomly sampled negatives based on part-of-speech (POS) perturbations. Specifically, we use spaCy\footnote{\url{https://spacy.io/}} to identify key primitives (\emph{verbs}, \emph{nouns}, \emph{adjectives}, \emph{adpositions}, and \emph{numerals}) and randomly replace them with alternatives drawn from the test prompt suite.
As shown in Table~\ref{tab:prompt}, incorporating negative prompts consistently improves performance over optimizing with the positive prompt alone. Overall Consistency increases slightly across all strategies, with the strongest gains achieved by LLM-constructed negatives. 
Specifically, GPT-4o negatives yield the best \emph{Color} and \emph{Spatial Relation} scores, suggesting that semantically targeted negatives provide more informative contrast than random perturbations. By contrast, POS-based random negatives offer little improvement and even reduce \emph{Spatial Relation}, indicating that arbitrary replacements are insufficient for effective calibration. Motion Smoothness remains essentially unchanged across variants, while Dynamic Degree changes only modestly, consistent with our design that negative-aware calibration primarily sharpens semantic and attribute alignment without degrading motion.

\begin{wraptable}{r}{0.54\columnwidth}
\vspace{-11pt}
\centering
\caption{Performance of different test-time scaling methods. 
}
\label{tab:tts}
\setlength{\tabcolsep}{3.5pt}
\renewcommand{\arraystretch}{1.08}
\footnotesize
\resizebox{0.50\columnwidth}{!}{%
\begin{tabular}{lcccccc}
\toprule
Models &
\makecell{Overall\\Consist.} &
\makecell{Multiple\\Objects} & Color &
\makecell{Spatial\\Relation.} &
\makecell{Motion\\Smooth.} &
\makecell{Dynamic\\Degree} \\
\midrule
T2V-Turbo (MS) & 27.40 & 54.50 & 89.90 & 46.62 & 95.51 & 70.83 \\
\midrule
+ NoisEasier & 32.05 & 76.62 & 94.38 & 48.93 & 95.29 & 71.39 \\
+ Best-of-\emph{n} sampling &  29.39 & 80.06 & 91.43 & 48.54 & 95.52 & 79.44 \\
+ PE (GPT-4o) & 28.70 & 71.95 & 87.47 & 47.58 & 95.89 & 70.56 \\
+ PE (Gemini-2.5 Flash) & 28.54 & 72.13 & 86.54 & 51.13 & 95.80 & 70.83 \\
\bottomrule
\end{tabular}%
}
\vspace{-15pt}
\end{wraptable}
\noindent \textbf{Comparison of test-time scaling methods.}
We further compare NoisEasier with two canonical test-time scaling (TTS) methods: Best-of-\emph{n} sampling (BoS) and prompt enhancement (PE). Under the same wall-clock time budget, BoS samples multiple seeds and selects the highest-reward video, while PE uses an LLM to iteratively rewrite the prompt (with fixed noise) and chooses the best result.
As shown in Table~\ref{tab:tts}, NoisEasier achieves the highest \emph{Overall Consistency} and \emph{Color} scores, indicating that gradient-based noise optimization is effective for semantic alignment and attribute binding. In contrast, BoS attains the best \emph{Dynamic Degree} and \emph{Multiple Objects}, suggesting that strong motion patterns and complex multi-object layouts are largely driven by the global structure induced by the initial noise. PE slightly improves \emph{Motion Smoothness} but substantially reduces \emph{Color} binding, implying that enriched prompts may distract the model from precise attribute alignment while stabilizing cross-frame predictions. Interestingly, Gemini-based PE achieves the best \emph{Spatial Relation} score, suggesting that some LLMs rewrite prompts with clearer spatial expressions. Overall, different TTS strategies favor different aspects of quality, and NoisEasier offers a complementary path for improving semantic and attribute-level alignment.

\vspace{2pt}

\begin{wraptable}{r}{0.54\columnwidth}
\vspace{-1.1cm}
\centering
\caption{Comparison with reward fine-tuning.}
\label{tab:rft}
\setlength{\tabcolsep}{3.5pt}
\renewcommand{\arraystretch}{1.08}
\footnotesize
\resizebox{0.50\columnwidth}{!}{%
\begin{tabular}{lcccccc}
\toprule
Models &
\makecell{Overall\\Consist.} &
\makecell{Multiple\\Objects} & Color &
\makecell{Spatial\\Relation.} &
\makecell{Motion\\Smooth.} &
\makecell{Dynamic\\Degree} \\
\midrule
T2V-Turbo (MS) & 27.40 & 54.50 & 89.90 & 46.62 & 95.51 & 70.83 \\
\hline
+ reward fine-tuning (RFT) & 28.58 & 73.49 & 90.95 & 51.16 & 98.65 & 74.72 \\
+ NoisEasier & 32.05 & 76.62 & 94.38 & 48.93 & 95.29 & 71.39 \\
+ RFT + NoisEasier & 31.91 & 84.56 & 94.72 & 54.71 & 98.65 & 77.50 \\
\bottomrule
\end{tabular}%
}
\vspace{-28pt}
\end{wraptable}

\noindent\textbf{Comparison with reward fine-tuning.}
Since NoisEasier optimizes rewards at inference time, we also examine whether incorporating the same rewards directly into the model parameters yields comparable benefits. To this end, we perform reward fine-tuning (RFT) using LoRA on the T2V-Turbo backbone under the same mixed reward objectives. We follow the default LoRA configuration of T2V-Turbo and fine-tune on the full VBench prompt suite ($\sim$1k prompts) for 10k steps, after which further training begins to exhibit overfitting.
As shown in Table~\ref{tab:rft}, both approaches improve over the baseline but with different strengths. RFT mainly improves motion-related metrics such as \emph{Motion Smoothness} and \emph{Dynamic Degree}, whereas NoisEasier yields larger gains on compositional dimensions including \emph{Multiple Objects} and \emph{Color}. Notably, applying NoisEasier on top of the RFT model further improves most metrics, indicating that reward signals are not fully amortized into the model parameters during offline tuning. This suggests that noise optimization provides a complementary \emph{test-time scaling} mechanism even when the backbone has already been reward-tuned.

\begin{table}[ht]
\centering
\caption{Inference Efficiency of 25-step test-time noise optimization.}
\label{tab:example}
\resizebox{\linewidth}{!}{
\begin{tabular}{l c c | l c c | l c c}
\toprule
 Model & Peak VRAM & Time & Model & Peak VRAM & Time & Model & Peak VRAM & Time \\
\midrule
MS & 8.23 GB & 1.04s & VC2 & 8.76 GB & 1.41s & Ani-LCM & 8.78 GB  & 2.46s \\
+Init. & 22.93 GB & 44.63s & +Init. & 33.24 GB & 115.19s & +Init. & 40.53 GB & 176.71s \\
+Full. & 22.95 GB & 45.28s & +Full. & 33.27 GB & 115.63s & +Full. & 40.53 GB & 177.86s \\
\bottomrule
\end{tabular}
}
\vspace{-10pt}
\end{table}

\subsection{Efficiency Analysis}

\noindent\textbf{Inference Efficiency.}
Table~\ref{tab:example} reports the peak memory and runtime of 25-step optimization under identical settings on a single RTX~6000 Ada GPU.
Compared with optimizing only the initial latent, full trajectory optimization incurs \emph{negligible} additional overhead, increasing runtime by less than $1\%$ while leaving peak memory virtually unchanged.
This is because both methods backpropagate through the same denoising computation graph; our method only introduces additional optimized noise variables without requiring extra forward or backward passes.
The dominant computational cost instead comes from multi-step backpropagation through the generator and differentiable reward models.
On a single H100 GPU, the runtime can be further reduced to 31.25\,s (MS), 72.92\,s (VC2), and 105.37\,s (AnimateLCM).

\vspace{2pt}

\begin{wrapfigure}{r}{0.5\textwidth}
    \centering
    \vspace{-20pt}
    \includegraphics[width=\linewidth]{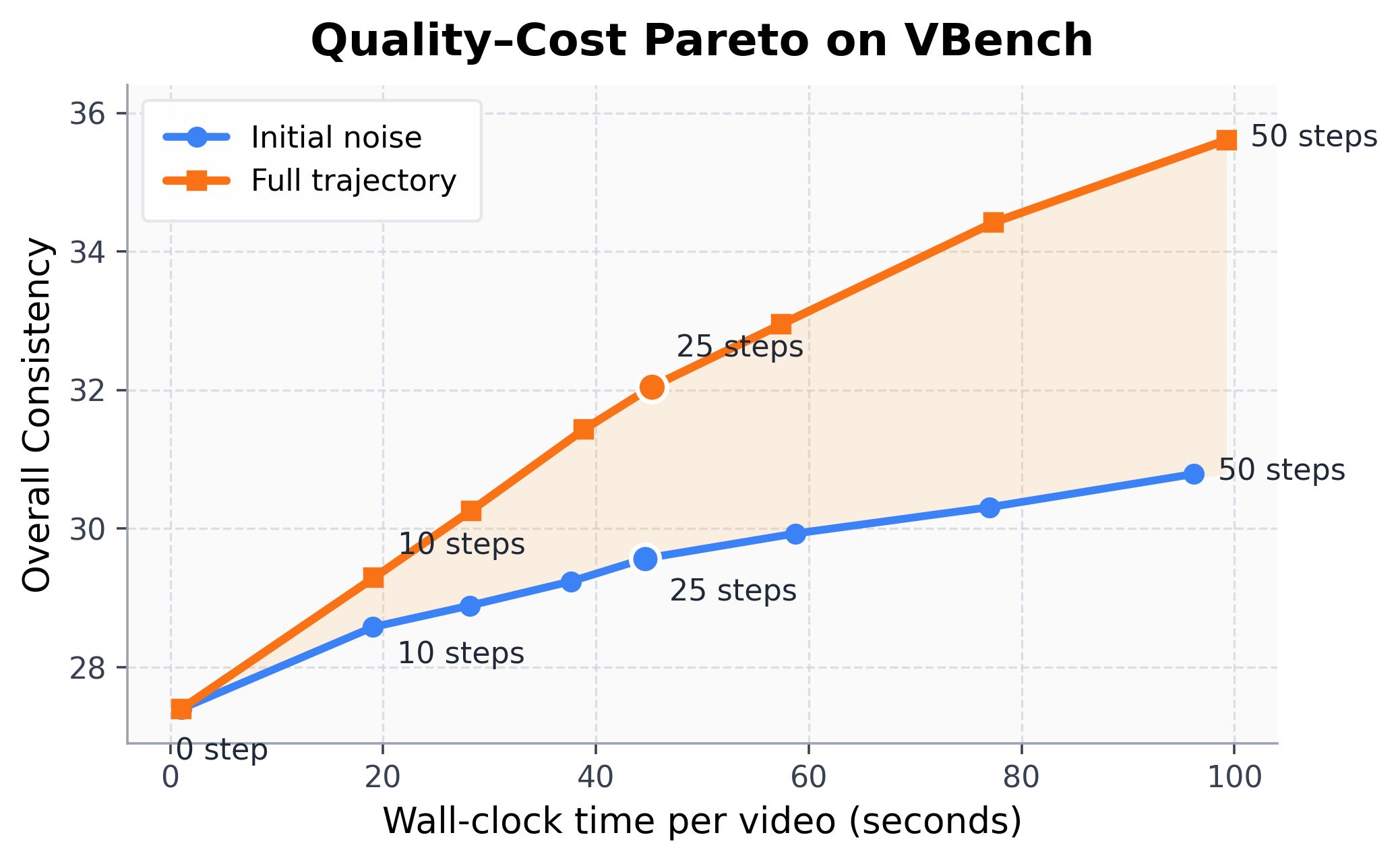} 
    \caption{Quality-Cost Pareto on VBench.}
    \label{fig:pareto}
    \vspace{-20pt}
\end{wrapfigure}
\noindent\textbf{Quality–Cost Trade-off.}
Figure~\ref{fig:pareto} compares the quality--cost trade-off between initial-noise and full-trajectory optimization on the T2V-Turbo (MS) backbone using an RTX 6000 Ada GPU, with the \emph{Overall Consistency} dimension of VBench as the evaluation metric. As the optimization budget increases, both methods improve semantic alignment, but full-trajectory optimization consistently traces a superior Pareto frontier. Notably, the advantage becomes more pronounced with additional optimization steps, indicating that jointly refining all intermediate perturbations utilizes the additional computation more effectively than optimizing only the initial latent. We adopt 25 optimization steps as the default setting, providing a favorable balance between quality and inference cost.

\subsection{Human Evaluation.}

\begin{wrapfigure}{r}{0.52\columnwidth}
\vspace{-20pt}
\centering
\includegraphics[width=0.50\columnwidth]{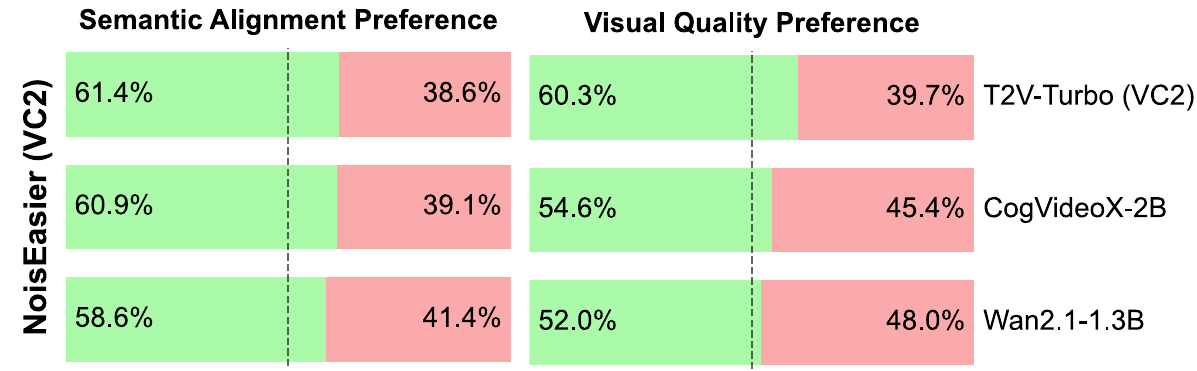}
\caption{User study on T2V-CompBench.}
\label{fig:user}
\vspace{-10pt}
\end{wrapfigure}
We conduct a user study to further assess the practical benefits of NoisEasier. We uniformly sample 10 prompts from each of the seven T2V-CompBench dimensions, resulting in 70 test prompts in total. We compare NoisEasier against three baselines: T2V-Turbo (VC2), CogVideoX-2B~\cite{yang2024cogvideox}, and Wan2.1-1.3B~\cite{wan2025wan}. For each prompt, we use the same random seed across models. Each comparison is judged by five Amazon Mechanical Turk workers, who select the candidate with better semantic alignment and visual quality.
As shown in Figure~\ref{fig:user}, NoisEasier achieves a 61.4\% win rate over its backbone, indicating that test-time noise optimization improves semantic faithfulness. 
Moreover, NoisEasier remains competitive against stronger T2V baselines, receiving around 61\% preference on alignment and 52\% on visual quality compared to CogVideoX-2B and Wan2.1-1.3B, despite operating as a lightweight add-on rather than retraining a larger T2V model.

\begin{figure}[t]
  \centering
  \includegraphics[width=\textwidth]{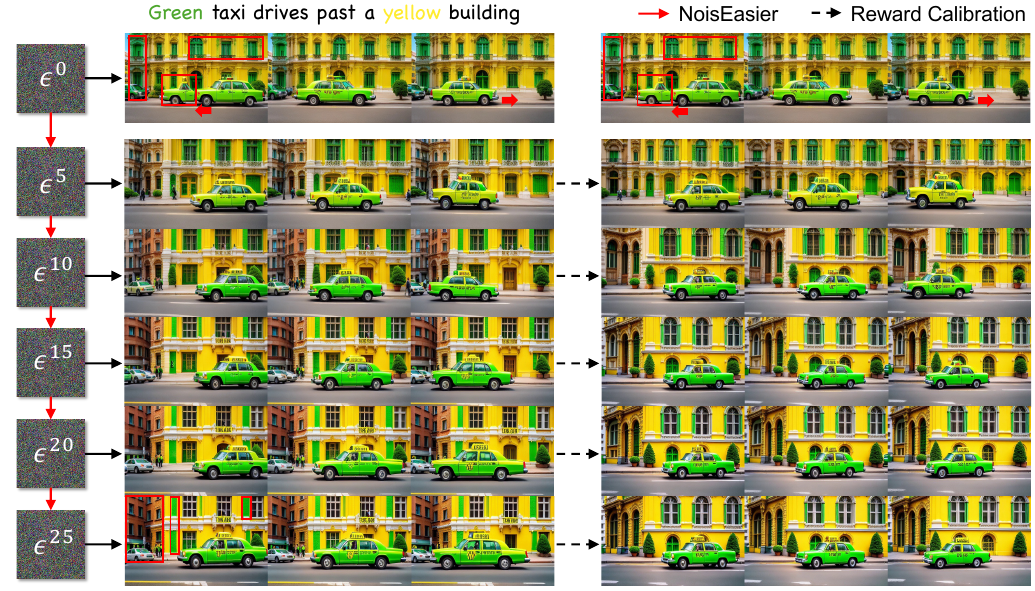}
  \caption{Effect of negative-aware reward calibration at different noise optimization steps.}
  \label{fig:case}
\end{figure}

\subsection{Qualitative Analysis}
We present qualitative comparisons in Figure~\ref{fig:case} to illustrate the effect of our optimization framework. The baseline captures the rough semantics of the prompt but exhibits compositional errors such as incorrect color binding and inconsistent object motion. Noise optimization progressively improves both spatial and temporal coherence: the taxi becomes a coherent object with stable right-to-left motion, and attribute separation becomes clearer, although minor artifacts remain. With negative-aware reward calibration, semantic disentanglement is further enhanced, i.e., the facade becomes consistently yellow while the green taxi is preserved with minimal leakage. 
Additional examples are provided in Figure~\ref{fig:head} and the supplementary material.

\begin{figure}[ht!]
    \centering
    \includegraphics[width=\linewidth]{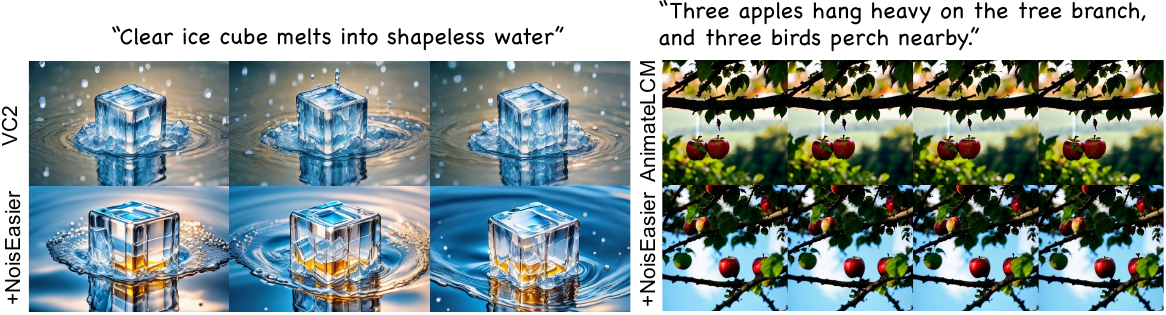}
    \caption{Two failure cases of NoisEasier: temporal state transition (left) and object counting (right).}
    \label{fig:fail}
\end{figure}

While effective, we also observe two common failure modes, as shown in Figure~\ref{fig:fail}.
First, although NoisEasier improves visual fidelity and semantic alignment, it still struggles with \emph{temporal state transitions}. In the melting-ice example, the generated video preserves the appearance of the ice cube but fails to capture its gradual transition into water. We attribute this limitation to existing reward models, which primarily assess frame-level semantics and provide only weak supervision for long-term temporal evolution.
Second, NoisEasier remains challenged by \emph{fine-grained object counting}. In the example on the right, the model fails to satisfy the exact numeracy constraint despite improving overall semantic consistency. This suggests that current VLM-based rewards are insufficiently sensitive to precise object counting and complex compositional reasoning.

These observations indicate that the performance of test-time optimization is fundamentally bounded by the quality of the underlying reward models. Future work may benefit from incorporating structured reward signals, such as object detection, visual grounding, or temporal-state verification models, to provide more reliable supervision for compositional video generation.
\section{Conclusion}
\label{sec:conclusion}


In this work, we propose NoisEasier, a test-time scaling framework that improves text-to-video generation through differentiable reward-guided noise optimization without model fine-tuning. By combining efficient optimization with robust reward supervision, NoisEasier consistently enhances compositional alignment while preserving visual fidelity. We further show that jointly optimizing intermediate noises provides more effective control over the denoising trajectory than optimizing only the initial latent, yielding faster reward convergence and stronger compositional alignment with negligible additional computational cost. Moreover, NoisEasier complements offline reward fine-tuning, demonstrating that test-time optimization remains effective even after model alignment. Overall, our findings establish noise optimization as a practical test-time scaling paradigm for improving alignment in video generative models.
\section*{Acknowledgements}
\label{sec:ack}

This work was partially supported by the Office of Naval Research (ONR) grant (N00014-23-1-2417), and the Army Research Office (ARO) grant (W911NF-24-1-0385). Any opinions, findings, and conclusions or recommendations expressed in this material are those of the authors and do not necessarily reflect the views of ONR or ARO.




%

{
\bibliographystyle{splncs04}
\bibliography{main}
}
\clearpage
\setcounter{page}{1}
\setcounter{section}{0}
\maketitlesupplementary

\newcommand{\cmark}{\textcolor{green!70!black}{\ding{51}}} 
\newcommand{\xmark}{\textcolor{red}{\ding{55}}} 

\renewcommand{\thesection}{\Alph{section}}

\section{More Implementation Details}
\label{sec:more_implentation}

Considering that different reward models produce outputs on different scales, we assign them separate weights $\alpha$ in Eq.~\ref{eq:update-rule}. Specifically, HPSv2 and ViCLIP measure cosine similarity between text and video, with most values falling in the range of $[0.2, 0.4]$, so their weights are set to $2$. The raw scores of ImageReward lie within $[-2, +2]$; we normalize them to $[0, 1]$ using min-max scaling to ensure non-negative rewards, and assign a weight of 1. For the motion reward, both the dynamic and smoothness terms are in $[0, 1]$, with $\lambda=0.5$ and an overall weight of $1$. We use GPT-4o to generate 8 negative samples per prompt. For negative-aware reward calibration, we set $m=0.02$ and $\tau=0.01$ for HPSv2, and $m=\tau=0.05$ for ViCLIP. We keep the original ImageReward since it is not based on video-text contrastive pretraining, and this also helps save memory and maintain inference speed. On both benchmarks, we empirically apply the above default settings for the two baseline models, which already demonstrate strong performance. Our code will be open-source upon acceptance.

\begin{figure*}[h]
  \centering
  \includegraphics[width=0.85\textwidth]{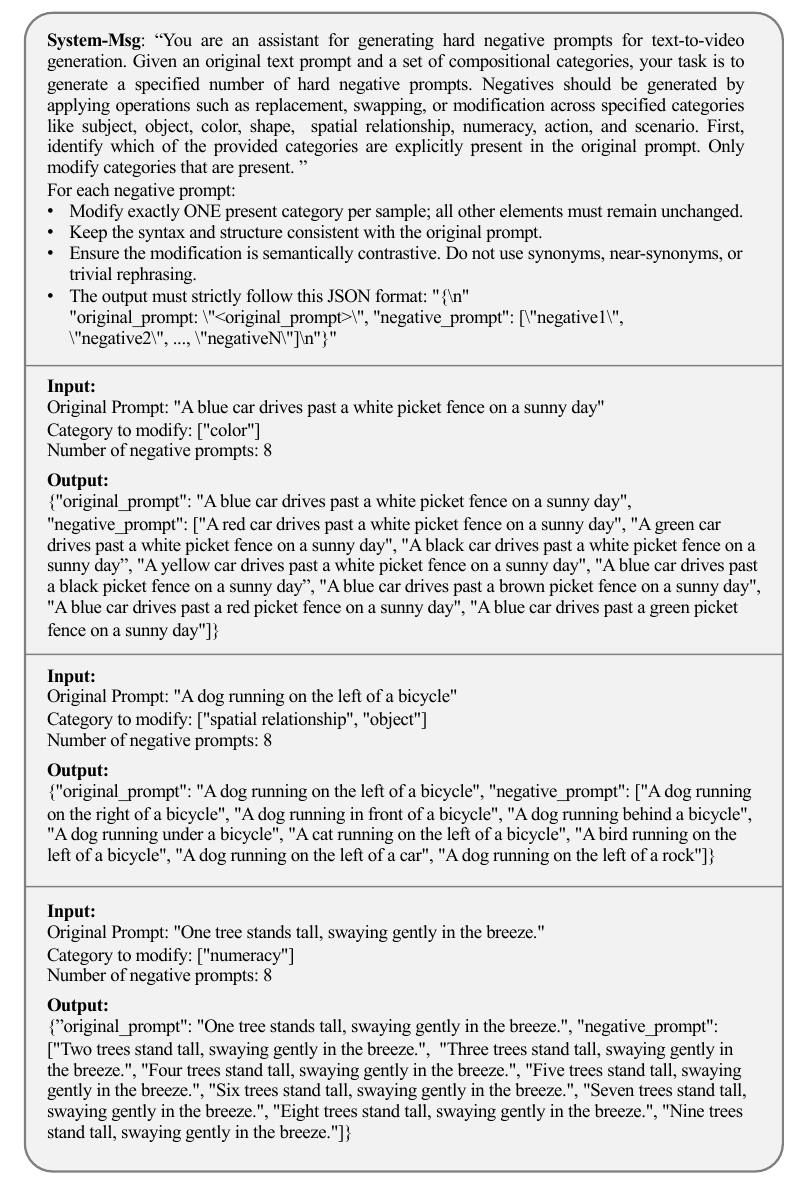}
  \caption{Prompt template and generated negative samples of GPT-4o.}
  \label{fig:prompt-template}
\end{figure*}

To construct hard negatives for reward calibration, we designed an automated pipeline using the OpenAI API. Figure~\ref{fig:prompt-template} illustrates the prompt template along with several generated examples. Given an input prompt from a specific dimension, the pipeline iterates through each prompt and queries GPT-4o with explicit instructions to generate syntactically similar but semantically different variants. Negatives are produced by modifying only one category at a time (\eg, subject, object, or action), ensuring that changes reflect clear semantic shifts rather than trivial rephrasings or synonyms. For each prompt, the model returns a fixed number of negatives (eight by default) in strict JSON format, which are automatically parsed, validated, and saved into a consolidated JSON file. This process yields a large pool of structured negatives that are consistent in format and directly usable for contrastive learning and evaluation.

\section{More Ablation Studies}
\label{app:ablation}

\noindent \textbf{Different Combination of Reward Models.}
In this section, we present all the combinations of the reward models in Table~\ref{tab:full-abl}. Among single-reward settings, ViCLIP delivers the strongest gains in overall consistency and color fidelity, highlighting its strength in semantic alignment and global appearance, whereas ImageReward performs best on multiple objects and spatial relations, reflecting its training on fine-grained visual-text cues. The motion reward is more distinctive: although it trails others on appearance-related metrics, it substantially boosts dynamic degree and smoothness, underscoring its importance in preventing static generations. When combining rewards, synergies are clear: pairing the motion reward with ViCLIP or ImageReward strikes a strong balance between temporal dynamics and spatial fidelity, while conflicts also appear; for instance, HPSv2 with other rewards can suppress motion expressiveness even as it improves color quality. Notably, the best three-reward combinations (\eg, ImageReward + ViCLIP + Motion) nearly match the four-reward setting.
Overall, the results show that different reward models specialize in complementary dimensions, and that careful combinations are crucial to balancing fidelity, compositionality, and temporal quality.

\begin{table*}[t]
\centering
\caption{All Combinations of reward models applied to NoisEasier (MS) on VBench. The best score for each number of reward models is highlighted in \colorbox{mygreen}{green}, and the second-best in \colorbox{myyellow}{yellow}. * indicates integrating negative-aware reward calibration.}
\setlength{\tabcolsep}{6pt}%
\scriptsize
\renewcommand{\arraystretch}{1.05}
\begin{adjustbox}{width=0.98\textwidth}
\begin{tabular}{cccccccccc}
\toprule
HPSv2 & ImgReward & ViCLIP & Motion  &
\makecell{Overall\\Consist.} &
\makecell{Multiple\\Objects} & Color & 
\makecell{Spatial\\Relation.} & 
\makecell{Motion\\Smooth.} & 
\makecell{Dynamic\\Degree} \\
\midrule
 \xmark  & \xmark  & \xmark  & \xmark  & 27.40 & 54.50 & 89.90 & 46.62 & 95.51 & 70.83 \\
\midrule
\cmark & \xmark  & \xmark  & \xmark  & 27.85 & 64.66 & 90.20 & \cellcolor{myyellow}46.21 & 95.19 & 54.72 \\
\xmark  & \cmark & \xmark  & \xmark  & \cellcolor{myyellow}28.24 & \cellcolor{mygreen}73.78 & \cellcolor{myyellow}90.27 & \cellcolor{mygreen}47.16 & \cellcolor{myyellow}95.36 & 61.11 \\
\xmark  & \xmark  & \cmark & \xmark  & \cellcolor{mygreen}36.21 & \cellcolor{myyellow}68.05 & \cellcolor{mygreen}91.19 & 41.92 & 94.98 & \cellcolor{myyellow}63.89 \\
\xmark  & \xmark  & \xmark  & \cmark  & 26.05 & 47.68 & 87.32 & 42.07 & \cellcolor{mygreen}95.59 & \cellcolor{mygreen}85.56 \\
\midrule
\cmark & \cmark  & \xmark  & \xmark  & 28.40 & \cellcolor{mygreen}74.02 & 89.35 & \cellcolor{mygreen}47.18 & \cellcolor{mygreen}95.53 & 55.28 \\
\cmark  & \xmark & \cmark  & \xmark  & \cellcolor{myyellow}34.57 & 70.78 & \cellcolor{mygreen}92.63 & 43.87 & 94.77 & 58.89 \\
\cmark  & \xmark  &  \xmark & \cmark  & 27.28 & 58.28 & 89.31 & 44.09 & \cellcolor{myyellow}95.52 & \cellcolor{myyellow}76.67 \\
\xmark  & \cmark  & \cmark  & \xmark  & 32.31 & 71.75 & \cellcolor{myyellow}90.69 & \cellcolor{myyellow}46.89 & 95.31 & 62.78 \\
\xmark  & \cmark  & \xmark  & \cmark  & 28.03 & \cellcolor{myyellow}73.26 & 89.34 & 44.43 & 95.43 & 74.72 \\
\xmark  & \xmark  & \cmark  & \cmark  & \cellcolor{mygreen}34.61 & 64.04 & 89.28 & 43.53 & 94.95 & \cellcolor{mygreen}80.28 \\
\midrule
\cmark & \cmark  & \cmark  & \xmark  & 31.95 &  \cellcolor{myyellow}74.09 & \cellcolor{myyellow}92.01 & 46.78 & 95.30 & 59.44 \\
\cmark  & \cmark & \xmark  & \cmark  & 28.20 &  73.16 & 91.50 &  46.36 &  \cellcolor{mygreen}95.38 & 75.28 \\
\cmark  & \xmark  &  \cmark & \cmark  & \cellcolor{mygreen}33.46 & 71.39 & 91.54 &  \cellcolor{myyellow}47.26 &  95.19 &  \cellcolor{mygreen}76.11 \\
\xmark  & \cmark  & \cmark  & \cmark  & \cellcolor{myyellow}31.96 & \cellcolor{mygreen}74.82 &  \cellcolor{mygreen}92.22 &  \cellcolor{mygreen}47.49 & \cellcolor{myyellow}95.35 & \cellcolor{myyellow}75.56  \\
\midrule
\cmark & \cmark & \cmark & \cmark & 31.74 & 75.82 & 93.29 & 47.95 & 95.38 & 73.33 \\
\cmark* & \cmark & \cmark* & \cmark & 32.05 & 76.62 & 94.38 & 48.93 & 95.29 & 71.39 \\
\bottomrule
\end{tabular}
\end{adjustbox}
\label{tab:full-abl}
\end{table*}

\noindent \textbf{Weight Sensitivity of Reward Models.}
As discussed in Sec.,\ref{sec:more_implentation}, different reward models operate on different numerical scales, so we assign separate weights $\alpha$ to HPSv2, ImageReward, ViCLIP, and the motion reward. Among the evaluated dimensions, Overall Consistency mainly reflects video--text semantic alignment, whereas the remaining dimensions capture object- and motion-level properties. We observe that semantic alignment varies only moderately across settings, and most metrics remain within a narrow range. One exception is Dynamic Degree, which shows higher variance because VBench uses a RAFT-based binary test (above or below a motion threshold), so small motion changes can flip the score. In contrast, Motion Smoothness exhibits very low variance, since it is computed from interpolation error and most samples already have reasonably smooth motion. As a result, changing the reward weights has limited effect on this metric.
In terms of trade-offs, placing greater weight on HPSv2 and ViCLIP slightly reduces Dynamic Degree, whereas emphasizing ImageReward and motion improves Dynamic Degree at some cost to Spatial Relation. Our final configuration strikes a balanced compromise, achieving competitive semantic alignment and strong object- and motion-related scores, while overall variation across settings remains modest.

\begin{table*}[t]
\centering
\caption{Performance of NoisEasier (MS) on VBench under different reward weight combinations. The \textbf{best} performance across all combinations is shown in bold, while the \textcolor{shade}{worst} performance is shaded in gray.}
\setlength{\tabcolsep}{6pt}%
\scriptsize
\renewcommand{\arraystretch}{1.05}
\begin{adjustbox}{width=0.98\textwidth}
\begin{tabular}{cccccccccc}
\toprule
HPSv2 & ImgReward & ViCLIP & Motion  &
\makecell{Overall\\Consist.} &
\makecell{Multiple\\Objects} & Color & 
\makecell{Spatial\\Relation.} & 
\makecell{Motion\\Smooth.} & 
\makecell{Dynamic\\Degree} \\
\midrule
 0.0  & 0.0  & 0.0  & 0.0  & 27.40 & 54.50 & 89.90 & 46.62 & 95.51 & 70.83 \\
\midrule
1.0 & 1.0 & 1.0 & 1.0  & 31.16 & 75.08 & \textcolor{shade}{91.16} & 46.27 & \textbf{95.55} & 70.56 \\
1.0 & 2.0 & 1.0 & 2.0  & 30.10 & 73.11 & 91.53 & 48.95 & 95.43 & 70.83 \\
5.0 & 1.0 & 5.0 & 1.0 & \textbf{32.76} & 75.43 & 92.88 & 47.64 & 95.35 & \textcolor{shade}{65.00} \\
2.0 & 2.0 & 1.0 & 1.0  & \textcolor{shade}{29.91} & 75.06 & 93.01 & \textbf{49.95} & 95.48 & 70.00 \\
2.0 & 1.0 & 1.0 & 2.0  & 30.55 & \textcolor{shade}{71.07} & 93.56 & 47.44 & 95.45 & 74.72 \\
1.0 & 2.0 & 2.0 & 1.0  & 31.45 & 76.25 & 93.06 & 46.01 & \textcolor{shade}{95.25} & 70.28 \\
1.0 & 1.0 & 2.0 & 2.0  & 31.90 & 72.01 & 93.15 & \textcolor{shade}{44.86} & 95.33 & \textbf{77.50} \\
2.0  & 1.0  & 2.0  & 1.0  & 32.05 & \textbf{76.62} & \textbf{94.38} & 48.93 & 95.29 & 71.39 \\
\bottomrule
\end{tabular}
\end{adjustbox}
\label{tab:weight}
\end{table*}

\begin{table}[ht]
\centering
\caption{Initial noise \emph{vs.} Full trajectory optimization on T2V-CompBench.}
\label{tab:compbench-abl}
\renewcommand{\arraystretch}{1.18}
\resizebox{\textwidth}{!}{%
\begin{tabular}{llllllll}
\toprule
Models &
\makecell{Consist\\ Attr.} &
\makecell{Dynamic \\Attr.} &
Spatial &
Motion &
Action &
Interaction &
Numeracy \\
\midrule
AnimateLCM+Init. & 0.7326 & 0.0121 & 0.4899 & 0.2291 & 0.4027 & 0.4469 & 0.2503 \\
\rowcolor{gray!15}
AnimateLCM+Full. & 0.7807 & 0.0183 & 0.5113 & 0.2281 & 0.4362 & 0.5365 & 0.2567 \\
\midrule
T2V-Turbo (MS)+Init.  & 0.8343 & 0.0094 & 0.5262 & 0.2370 & 0.5989 & 0.6243 & 0.3641 \\
\rowcolor{gray!15}
T2V-Turbo (MS)+Full. & 0.8375 & 0.0147 & 0.5259 & 0.2502 & 0.6477 & 0.6743 & 0.3983 \\
\midrule
T2V-Turbo (VC2)+Init. & 0.8524 & 0.0092 & 0.5434 & 0.2207 & 0.6961 & 0.7765 & 0.3781 \\
\rowcolor{gray!15}
T2V-Turbo (VC2)+Full. & 0.8737 & 0.0097 & 0.5660 & 0.2343 & 0.7524 & 0.7915 & 0.3753 \\
\bottomrule
\end{tabular}%
}
\end{table}

\section{Additional Results on Modern Video Generators}
Since NoisEasier performs iterative gradient-based optimization, directly applying it to large diffusion backbones with long denoising chains, e.g., Wan2.1-14B~\cite{wan2025wan} or HunyuanVideo~\cite{kong2024hunyuanvideo}, remains computationally prohibitive. For example, generating a single 5-second video with Wan2.1-14B requires approximately 56 minutes and 69~GB peak VRAM on a single A100 GPU, making repeated forward/backward optimization impractical. 
To further evaluate the generality of NoisEasier, we additionally apply our framework to two recent distilled video generators, FastWan\,2.1-1.3B\footnote{\url{https://github.com/hao-ai-lab/fastvideo}} (81 frames, 3-step sampling) and LTX-Video-2B-distilled~\cite{hacohen2024ltx} (121 frames, 10-step sampling). 

\begin{table}[ht]
\centering
\begin{minipage}{0.48\linewidth}
\centering
\captionof{table}{VBench Evaluation.}
\resizebox{\linewidth}{!}{
\begin{tabular}{lccc}
\toprule
Model & Consist. & Object & Color \\
\midrule
FastWan\,2.1-1.3B & 21.49 & 58.43 & 87.77 \\
+ NoisEasier   & 23.71 & 68.09 & 90.01 \\
\bottomrule
\end{tabular}
}
\vspace{-0.3em}
\label{tab:fastwan}
\end{minipage}
\hfill
\begin{minipage}{0.48\linewidth}
\centering
\captionof{table}{T2V-CompBench.}
\resizebox{\linewidth}{!}{
\begin{tabular}{lccc}
\toprule
Model & Consist. & Action & Interaction \\
\midrule
LTXV-2B-distilled & 0.7702 & 0.3631 & 0.4135 \\
+ NoisEasier   & 0.8081 & 0.4230 & 0.4931 \\
\bottomrule
\end{tabular}
}
\vspace{-0.3em}
\label{tab:ltxv}
\end{minipage}
\vspace{-0.5em}
\end{table}

\begin{center}
    \includegraphics[width=\linewidth]{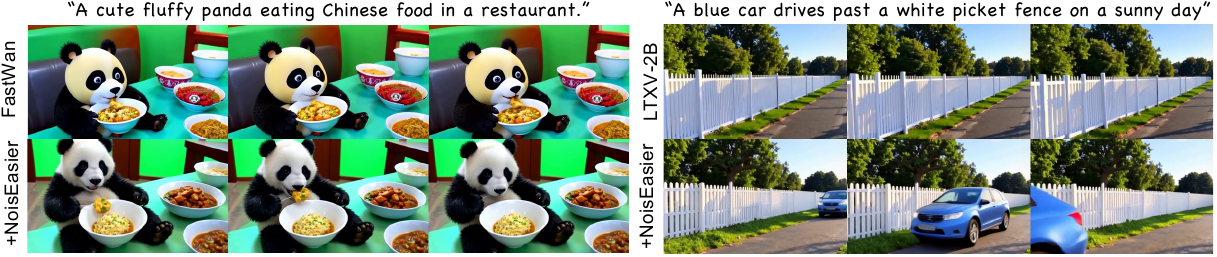}
    \captionof{figure}{Qualitative results on two modern fast generators.}
    \label{fig:rebuttal_new}
\end{center}
\vspace{0.5em}


Tables~\ref{tab:fastwan} and~\ref{tab:ltxv} show that NoisEasier consistently improves both FastWan and LTX-Video-2B-distilled under the default optimization setting. On FastWan, NoisEasier improves \emph{Overall Consistency} by 2.22\% while also yielding substantial gains in object recognition and color consistency. Similarly, on LTX-Video-2B-distilled, NoisEasier consistently improves overall consistency, action understanding, and object interaction. Figure~\ref{fig:rebuttal_new} further presents representative qualitative examples, demonstrating improved semantic alignment across diverse prompts. These preliminary results suggest that NoisEasier generalizes beyond the VCM backbones studied in the main paper and remains effective on modern distilled video generators.

\section{More Visualization Results}

\begin{figure}
  \centering
  \includegraphics[width=0.9\textwidth]{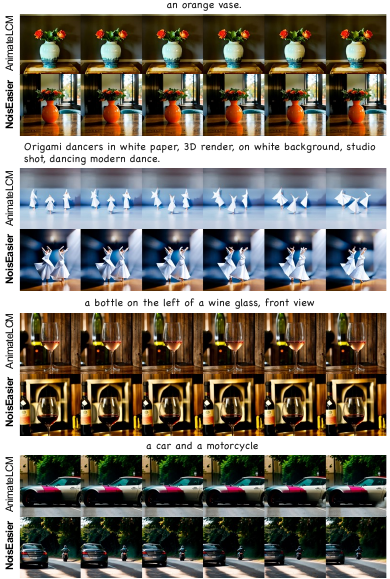}
  \caption{Qualitative comparison between AnimateLCM and NoisEasier on VBench.}
  \label{fig:animate_vbench}
\end{figure}

\begin{figure}
  \centering
  \includegraphics[width=0.9\textwidth]{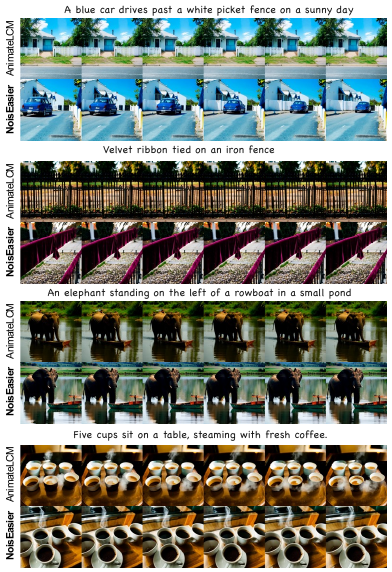}
  \caption{Qualitative comparison between AnimateLCM and NoisEasier on T2V-CompBench.}
  \label{fig:animate_compbench}
\end{figure}

\begin{figure}
  \centering
  \includegraphics[width=\textwidth]{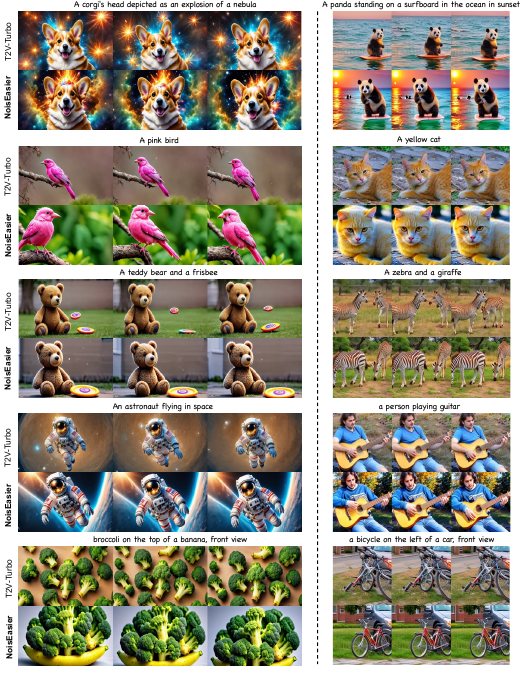}
  \caption{Qualitative comparison between baseline results and NoisEasier on VBench. \textbf{Left}: T2V-Turbo (VC2). \textbf{Right}: T2V-Turbo (MS).}
  \label{fig:appendix-vbench}
\end{figure}

\begin{figure}
  \centering
  \includegraphics[width=\textwidth]{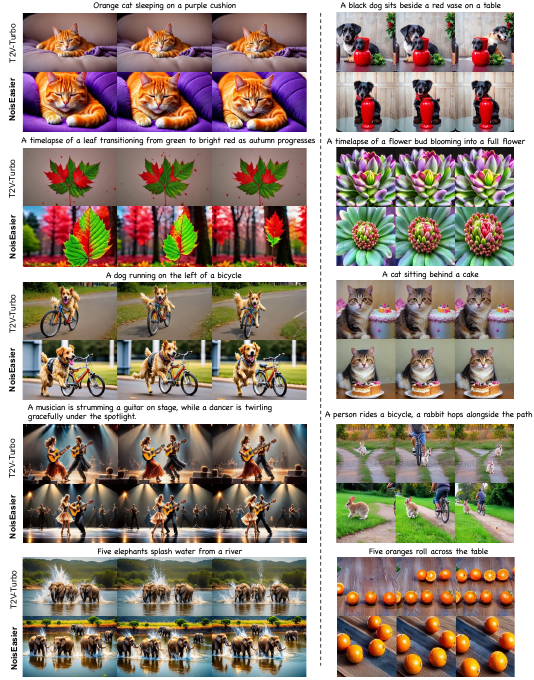}
  \caption{Qualitative comparison between baseline results and NoisEasier on T2V-CompBench. \textbf{Left}: T2V-Turbo (VC2). \textbf{Right}: T2V-Turbo (MS).}
  \label{fig:appendix-compbench}
\end{figure}

We provide more qualitative examples from Figure~\ref{fig:animate_vbench} to Figure~\ref{fig:appendix-compbench} to illustrate the effectiveness of our method. 

\section{Limitations and Future Work}
As a test-time optimization framework, NoisEasier inevitably introduces additional inference overhead compared with vanilla sampling, making it less suitable for latency-sensitive applications. Moreover, it primarily improves semantic alignment rather than the intrinsic visual quality of the backbone, which remains bounded by the underlying generator and imperfect reward proxies. Our study also focuses on short, single-scene video generation; extending noise refinement to longer videos, more complex narratives, or higher resolutions remains challenging due to weaker reward signals and increased optimization cost. Future work will explore stronger video-level rewards and more efficient optimization strategies for long-horizon video generation.

\end{document}